# Estimating Forces of Robotic Pouring Using a LSTM RNN

Author: Kyle Mott

*Abstract*—In machine learning, it is very important for a robot to be able to estimate dynamics from sequences of input data. This problem can be solved using a recurrent neural network. In this paper, we will discuss the preprocessing of 10 states of the dataset, then the use of a LSTM recurrent neural network to estimate one output state (dynamics) from the other 9 input states. We will discuss the architecture of the recurrent neural network, the data collection and preprocessing, the loss function, the results of the test data, and the discussion of changes that could improve the network. The results of this paper will be used for artificial intelligence research and identify the capabilities of a LSTM recurrent neural network architecture to estimate dynamics of a system.

## I. INTRODUCTION

Robotics are becoming more prominent and applicable in everyday life's tasks. For daily tasks, robots need to be able to perform manipulation to execute objectives. Recently, robotic pouring has been a popular research study for machine learning. For robotic pouring, the motion and dynamics of pouring varies with respect to time while most of the input parameters are constant. Because of this, estimating the motion and dynamics of pouring with respect to time can be difficult due to the inherent difficulty of estimating liquid dynamics.

Various techniques of robot pouring [1], [2] and [3] have been tried over the years. In [1], model learning to estimate the pouring container symmetric geometry has been tried. The robot would determine the volumetric flow rate from the estimated container geometry along with a time delay [1]. This type of model learning was tried on new containers not in the dataset quickly in a single attempt [1]. This type of research achieved under 5ml error between 20 to 45 second pours [1]. Likewise, in [2], collision-free trajectory algorithms have been tried for trajectory planning for pouring liquids. A fine-grained liquid simulator was used to guide the trajectory optimization and integrate them into planning framework [2]. Also, adaptive pouring techniques have been tried using different container sizes [3]. A two-stage teaching process was tried to pour liquids based on simulation [3].

This report focuses on using manipulation learning with a Recurrent neural network (RNN) to take inputs and hidden states from previous time steps to produce an output for the current time step. Recurrent neural networks are suitable for estimating the dynamics of robot pouring based on pouring due to the inherent capability of recurrent neural networks. A significant advantage recurrent neural networks have over convolutional neural networks is the input shape of the network can vary with the input data. Because of this, recurrent neural networks are suitable for input sequential time data. With Recurrent neural networks, a significant problem of the vanishing gradient and exploding gradient can occur during input of data in the model. To deal with this, we can use two types of models, LSTM and GRU.

Both Gated Recurrent Unit (GRU) and Long Short-Term Memory (LSTM) have the ability to keep memory from past activation rather than replacing the entire activation like a standard recurrent neural network would [4]. The likelihood of the vanishing gradient is significantly reduced because this ability allows features to be remembered through backpropagation through multiple nonlinearities [4]. Through the use of truncating the gradients, the likelihood of the exploding gradient occurring can be significantly reduced.

Gated Recurrent Unit (GRU) have been used in recurrent neural networks to deal with time sequential data and gradient problems. Different variants of the Gated Recurrent Unit have been tried in recurrent neural networks [5]. In [5], three different variants, GRU1, GRU2, and GRU3, were done in a comparative study. Gated Recurrent Units have been demonstrated to show superiority over traditional Tanh units [6]. Through this study, results clearly showed the significant advantages of gated units over traditional vanilla recurrent units of learning rates per epoch on validation runs [6].

Long Short-Term Memory (LSTM) have also been used in recurrent neural networks as commonly as Gated Recurrent Units. LSTMs have been effective at capturing long-term temporal dependencies [7]. In [7], study has been done to empirically compare various LSTM architecture structures. Another study [8] has been shown to show how LSTM can be an approach for tasks that require accurate measurement between time intervals. This study [8] recognizes the development of LSTM and shows how LSTM can solve nonlinear tasks.

In this project, we will discuss the use of LSTMs to estimate the dynamics of robotic pouring from nine inputs. The goal of this report is to deal with time varying sequences to indulge in the architectural design of the recurrent neural network and different structures to see how well graphically the dynamics of robotic pouring can be estimated from the input data.

## II. DATA COLLECTION AND PREPROCESSING

### A. Dataset

The dataset for this project includes 1307 motion sequences and corresponding weight measurements. Each sequence had varying time sequences with a maximum of 1099. The 10 features for each time sequence are as follows:

$$\theta(t) = Rotation\ Angle\ at\ time\ t\ (Degrees) \quad (1)$$
$$f(t) = Weight\ at\ time\ t\ (lbf) \quad (2)$$
$$f_{init} = Weight\ before\ pouring\ (lbf) \quad (3)$$
$$f_{empty} = Weight\ while\ cup\ is\ empty\ (lbf) \quad (4)$$
$$f_{final} = Weight\ after\ pouring\ (lbf) \quad (5)$$
$$d_{cup} = Diameter\ of\ receiving\ cup\ (mm) \quad (6)$$
$$h_{cup} = Height\ of\ receiving\ cup\ (mm) \quad (7)$$
$$d_{ctn} = Diameter\ of\ pouring\ cup\ (mm) \quad (8)$$
$$h_{ctn} = Height\ of\ pouring\ cup\ (mm) \quad (9)$$
$$\rho = Density\ of\ material/water\ (unitless) \quad (10)$$

Feature dimensions (1) and (2) vary with time while the other eight of the dimensions are held constant. The process of the data collection can be found in [9], [11], and [12].

### B. Data and Preprocessing

First, the 1307 motion sequences were parsed into a total input data of size (1307, 1099, 9) with inputs of features (1, 3, 4, 5, 6, 7, 8, 9, 10) and a total target data of size (1307, 1099, 1) with a target of feature (2). The data is then split into training by 80%, validation by 15%, and testing by 5%. The data sizes can be seen in the Table 1:

Table 1: Split Data Size

| Data Type | Size |
|---|---|
| Training Input | (1045, 1099, 9) |
| Training Target | (1045, 1099, 1) |
| Validation Input | (196, 1099, 9) |
| Validation Target | (196, 1099, 1) |
| Testing Input | (66, 1099, 9) |
| Testing Target | (66, 1099, 1) |

Each sequence was time varying but was padded with zeros to make each sequence have a time length of 1099. At first, we were worried these zeros would affect the estimated target function by the loss function considering the padded zeros at the end of the sequence into account. To prevent this, a custom loss function was created to mask the zeroes before loss is calculated. The input size remained (1099, 9), but the zeroes were masked as validation was calculated. The custom loss function yielded higher validation loss than without masking the zeroes; but, this is due to the zeroes contributing to the loss inflating the difference.

## III. METHODOLOGY

Several sequential LSTM networks were used to estimate the dynamics of the pouring system from the 9 inputs. For the first model, a single LSTM 16 layer with activation functions "Tanh" and recurrent activation functions "Sigmoid" was used as a base model to start training. The starting model can be seen in Figure 1.

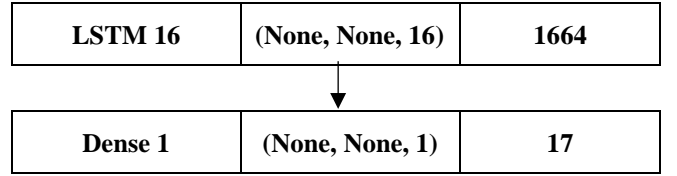

*Figure 1: Starting Model*

This starting model was used as a baseline for the LSTM architecture to come and more layers were added on to reduce the validation loss.

In the second model proposed, 4 additional layers of LSTM were added to the starting model with a dropout layer of 0.2. This model introduced a regularization technique of dropout while increasing the number of LSTM layers. The attempt here was to reduce the overall validation loss while the network not becoming susceptible to overfitting. The LSTM cells were kept to 16 for the layers. The second model can be seen in Figure 2.

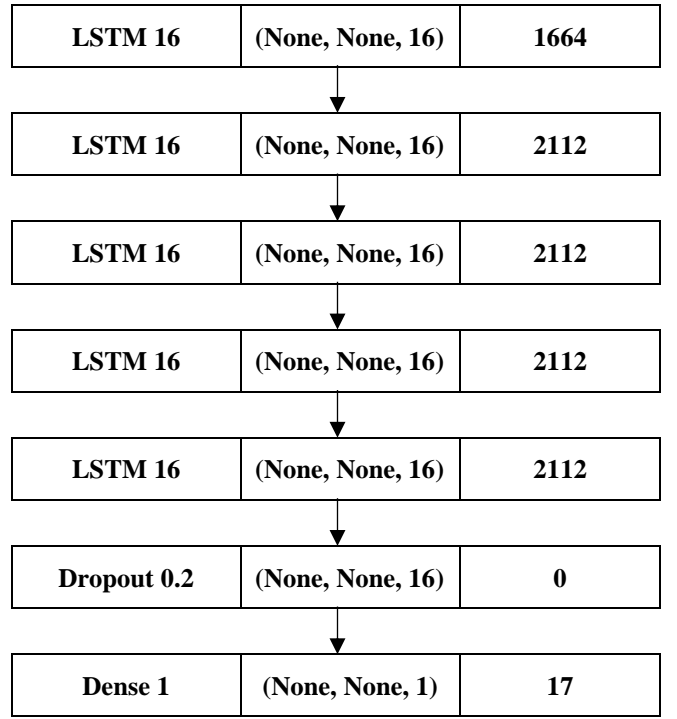

*Figure 2: Second Model*

The second model showed major improvement from the starting model drastically reducing the validation accuracy.

In the final model, additional layers were added on the second model. The additional layers added was one LSTM layer followed with another dropout of 0.15 and lastly an output dense layer of 1. The final model architecture can be seen in Figure 3.

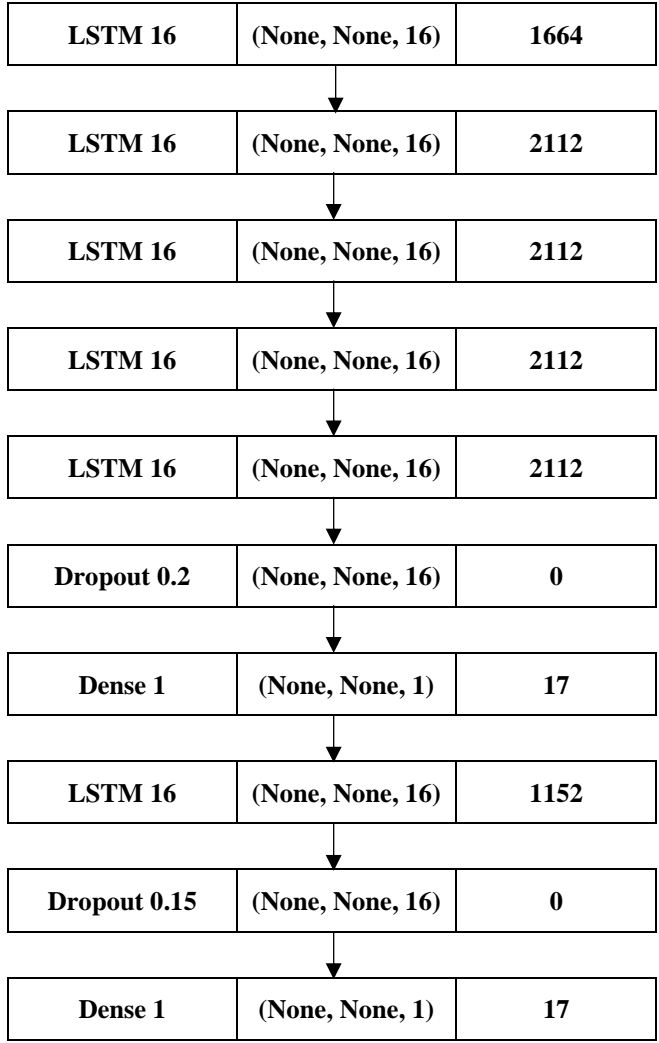

*Figure 3: Final Model*

In our final model, we added an additional LSTM layer and regularization of dropout after the model was fully connected.

The optimizer used in all the models was Adam with a learning rater of 0.0001. Adam was chosen because of its ability to converge quickly while traditionally performing better than most other optimizers.

The two loss functions used in the final model was Mean Square Error (MSE) and a custom loss function. Mean square error [10] is the most commonly used regression loss function. It is the sum of squared distances between the predicted values and the target value divided by the number of squared differences. The mean square error equation is shown in (11).

$$MSE = \frac{\sum_{i=1}^{n}(y_i - y_i^p)^2}{n} \tag{11}$$

The custom loss function was used to mask the zeroes at the end of each sequence to prevent the zeroes from being calculated in the loss. The custom loss function continued to use mean square error on the function after the zeroes were omitted.

## IV. EVALUATION AND RESULTS

### A. Testing Data Split from Initial Dataset

The testing dataset of 66 motion sequences (5%) was split from the original 1307 motion sequences and was used to verify the initial results. This testing data results was compared between models to verify results from architectural changes. The second model ended with very low validation accuracy at around 0.00586 by the tenth epoch. The Loss vs Epochs graph can be seen in Figure 4.

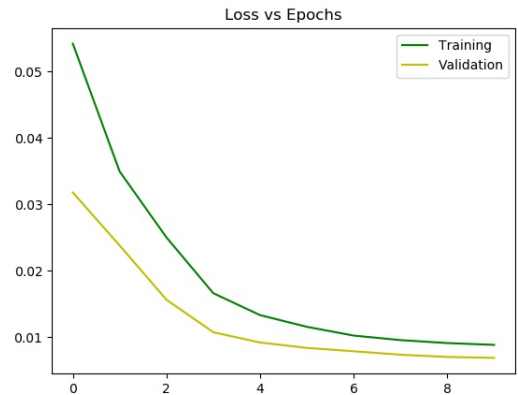

*Figure 4: Second Model, Loss vs Epochs*

While the second model's architecture showed very low validation accuracy, it predicted forces well in most lower magnitudes of forces but failed to predict forces in larger changes in magnitudes of forces. In Figure 5, the prediction closely matches the forces for this motion sequence.

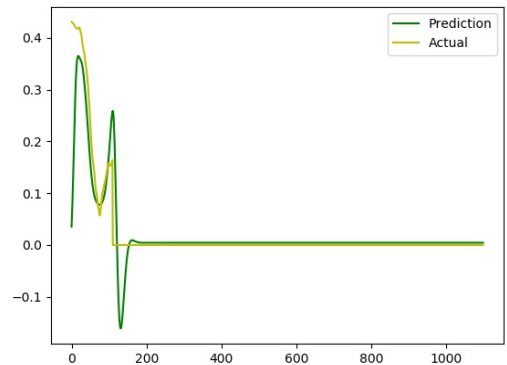

*Figure 5: Second Model, Force Estimation 1*

In Figure 6, the prediction struggles to capture the actual forces represented in the larger change in magnitude.

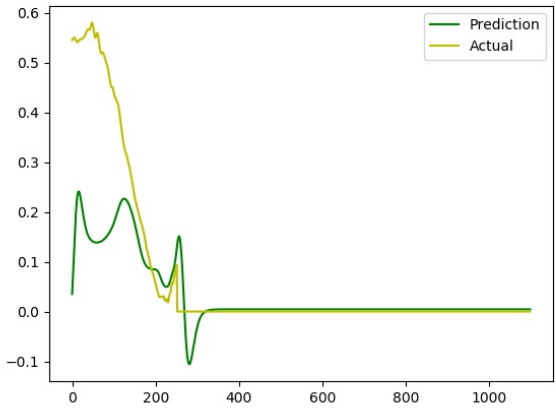

*Figure 6: Second Model, Force Estimation 2*

For the final model, the predictions estimated the forces much closer than the second model. The final model achieved a final validation accuracy of 0.00286. Nearly all the 66 test sample predictions closely matched the actual data.

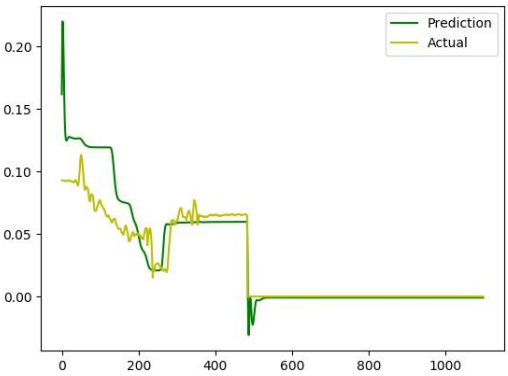

*Figure 7: Final Model, Force Estimation 1*

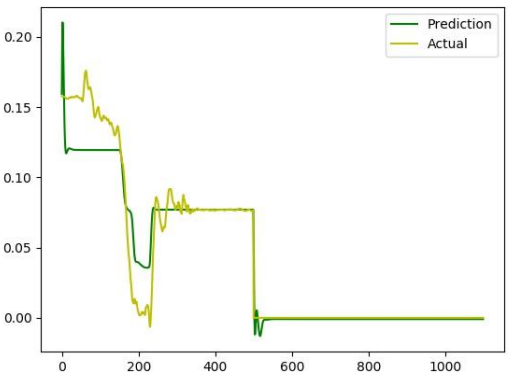

*Figure 8: Final Model, Force Estimation 2*

From Figures 7 and 8, the prediction closely matches the larger changes in magnitude of forces in which wasn't occuring in the second model.

### B. Unseen Test Data, 6 Plots

The unseen dataset contained 289 motion sequences with a different cup not contained in the dataset the model trained on. Also, the unseen test data was given in a different input shape than the first dataset. The unseen dataset had time length of 834 compared to the initial 1099 with the zeroes added to the end of sequences. Since the final model architecture can accept variable input shapes, there wasn't a problem adjusting to the unseen dataset. Of the 289 motion sequences in the unseen dataset, 6 sequences were chosen to analyze results. The 6 sequences chosen are indices 98, 27, 60, 248, 226, 177. Figures 9 – 14 show the plots of the actual vs prediction force estimation.

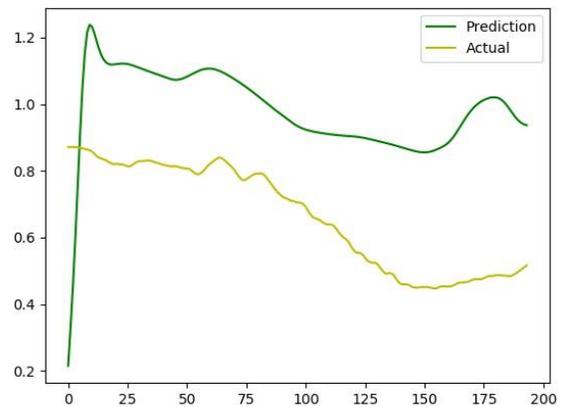

*Figure 9: Unseen Dataset, Motion Sequence Index 98*

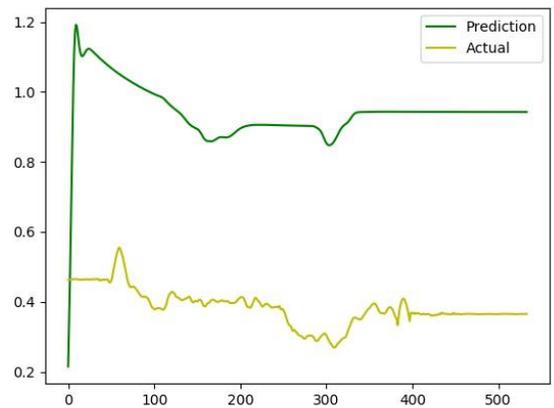

*Figure 10: Unseen Dataset, Motion Sequence Index 27*

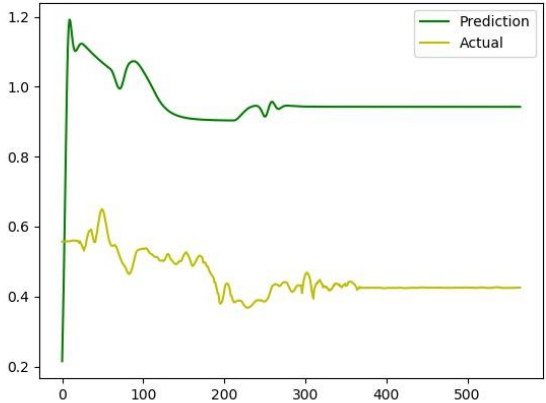

*Figure 11: Unseen Dataset, Motion Sequence Index 60*

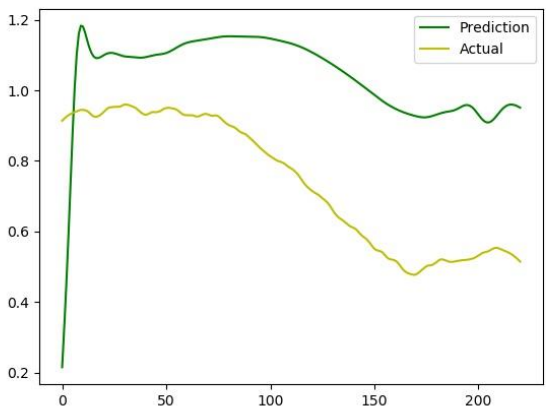

*Figure 12: Unseen Dataset, Motion Sequence Index 248*

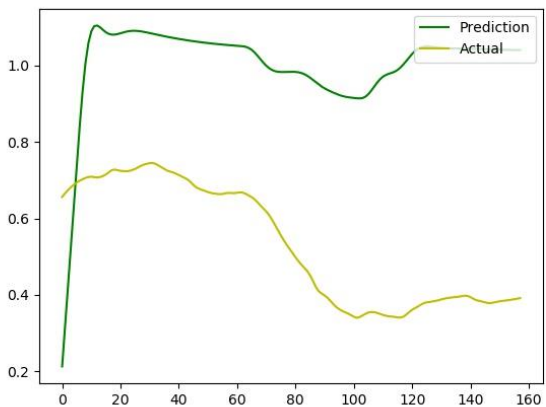

*Figure 13: Unseen Dataset, Motion Sequence Index 226*

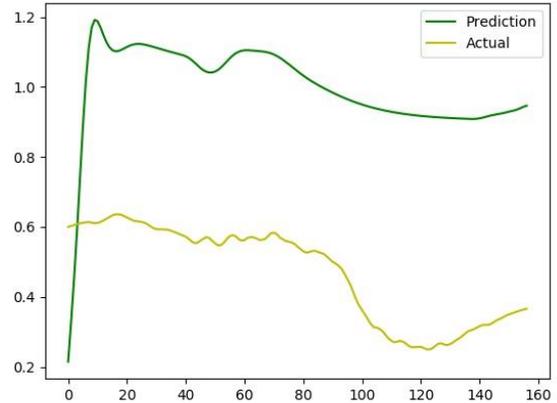

*Figure 14: Unseen Dataset, Motion Sequence Index 177*

From the graphs, we have seen that our model tries to predict the dynamics of the pouring from an unknown cup size. The prediction in all of the results seems to be a higher magnitude than the actual result. However, the prediction does closely match the slope of the actual forces. The reason for this could be the lack of normalization of the data. We can better estimate the dynamics by normalizing the data before inputting it into our model. Normalization wasn't done during this project on the constant inputs but for the future it could help generalize the data.

## V. CONCLUSION

Estimating dynamics from input states is just one of the main challenges faced in robotic pouring. For this challenge, a LSTM Recurrent neural network was used to perform training and test on unseen data. Through multiple iterations of architectures, our final model proved to give better results than preliminary models through the implementation of regularization and LSTM layer structure. For testing and validation split from the initial dataset, the final model captures the estimation well. For the unknown cup dataset, the model predicts the actual forces but struggles with precision. Normalization of the constant inputs could help with this problem. Also, it's possible that a different LSTM model architecture could have better results than the final model proposed in this project. For future work, there are many other techniques this model didn't indulge into that could help estimate the dynamics of unknown pouring cups.